\documentclass[10pt,twocolumn,letterpaper]{article}

\usepackage[pagenumbers]{cvpr} 

\usepackage{amssymb}
\usepackage{array} 
\usepackage{amsmath}
\usepackage{graphicx}
\usepackage{float}
\usepackage[linesnumbered,ruled,vlined,longend]{algorithm2e}

\definecolor{cvprblue}{rgb}{0.21,0.49,0.74}
\usepackage[pagebackref,breaklinks,colorlinks,allcolors=cvprblue]{hyperref}

\def\paperID{17} 
\def\confName{CVPR PBVS Workshop}
\def\confYear{2025}

\title{LangGas: Introducing Language in Selective Zero-Shot Background Subtraction for Semi-Transparent Gas Leak Detection with a New Dataset}

\author{
    Wenqi Marshall Guo$^{1,2}$ \quad
    Yiyang Du$^{2,3}$ \quad
    Shan Du$^{1,*}$ \\
    $^1$Department of CMPS, University of British Columbia, Canada \\
    $^2$Group of Methane Emission Observation \& Warning (MEOW) , Weathon Software, Canada \\
    $^3$Department of Computational Linguistics, University of British Columbia, Canada \\
    *Corresponding Author\\
    {\tt\small wg25r@student.ubc.ca, duyiyang@student.ubc.ca, shan.du@ubc.ca}
}

\begin{document}
\flushbottom
\maketitle
\begin{abstract}
Gas leakage poses a significant hazard that requires prevention. Traditionally, human inspection has been used for detection, a slow and labour-intensive process. Recent research has applied machine learning techniques to this problem, yet there remains a shortage of high-quality, publicly available datasets. This paper introduces a synthetic dataset, SimGas, featuring diverse backgrounds, interfering foreground objects, diverse leak locations, and precise segmentation ground truth. We propose a zero-shot method that combines background subtraction, zero-shot object detection, filtering, and segmentation to leverage this dataset. Experimental results indicate that our approach significantly outperforms baseline methods based solely on background subtraction and zero-shot object detection with segmentation, reaching an IoU of 69\%. We also present an analysis of various prompt configurations and threshold settings to provide deeper insights into the performance of our method. Finally, we qualitatively (because of the lack of ground truth) tested our performance on GasVid and reached decent results on the real-world dataset. The dataset, code, and full qualitative results are available at  \url{https://github.com/weathon/Lang-Gas}.
\end{abstract}

\section{Introduction}
\label{sec:intro}
Various organic gases are extensively used today across industry, both as fuels (for example, natural gas, methane) and starting materials for synthesis. However, methane emissions and other gases, particularly unintended leaks, are harmful and should be prevented. Methane has a significantly greater greenhouse impact than carbon dioxide (CO\(_2\)), exhibiting a heat-trapping potential over 28 times that of CO\(_2\) \cite{USEPA_2016}. Additionally, many hydrocarbon gases can be toxic to humans \cite{Curtis_Metheny_Sergent_2025}. In enclosed areas, leaked gases may pose a risk of hypoxia, endangering personnel, or causing fire and explosion hazards. For additional background, detailed justification, and the need for effective leak detection, readers may refer to several previously published papers \cite{gasvid,gasformer,iig}.
\begin{figure}
    \centering
    \includegraphics[width=\linewidth]{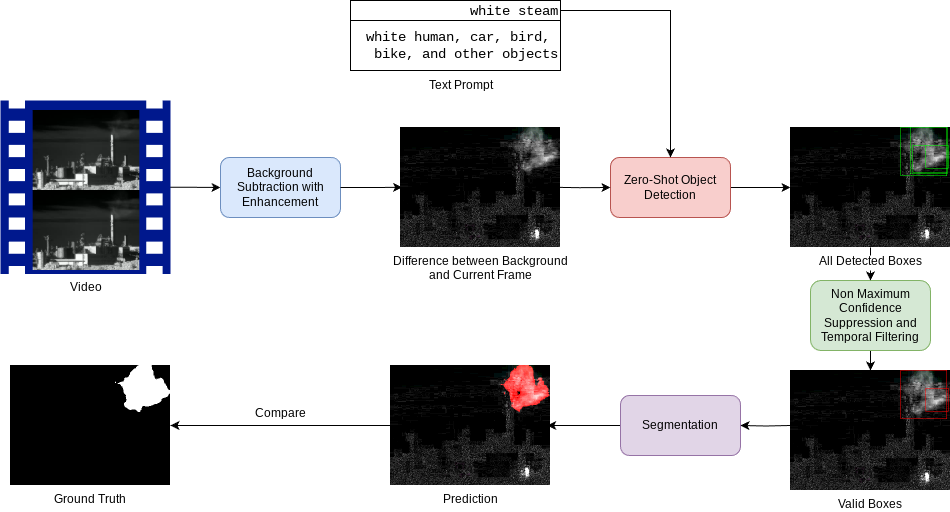}
    \caption{\textbf{Method overview: } Our method for gas leak detection involves background subtraction, zero-shot object detection, non-maximum suppression (NMS), temporal filtering, and segmentation. First, background subtraction is used to identify the moving parts in the video. Then, two text prompts (positive and negative prompts) are employed to guide a zero-shot object detector in detecting leaks. We use the prompt "white steam" because it is more commonly recognized than phrases explicitly mentioning gas leaks. NMS and temporal filtering are then applied to remove extra boxes and fix false positives or negatives based on past temporal information. Finally, a segmentation model—such as the Segment Anything Model 2 (SAM 2)—is used to convert the bounding boxes into segmentation masks.}
    \label{fig:pipeline}
\end{figure}

There are many works focusing on gas leak detection using computer vision; however, despite many computer vision algorithms being data-intensive, public datasets are very scarce. Three major datasets in this field are GasVid \cite{gasvid}, Gas-DB \cite{gasdb}, and the Industry Invisible Gas Dataset (IIG) \cite{iig}.

GasVid \cite{gasvid} is a dataset featuring controlled methane gas releases against a clear sky background, making it nearly ideal for foreground (leak) segmentation via background subtraction. Yet, this scenario rarely reflects real-world conditions. Moreover, GasVid lacks segmentation ground truth, providing only quantification classification that limits segmentation assessment to visual inspection. Gas-DB \cite{gasdb} includes segmentation but consists of still images instead of full-motion video. Although the images are sequential, they exhibit low continuity and short duration, making many video-based methods inapplicable. Both GasVid and Gas-DB rely on human-controlled releases, where leaks originate at the end of a releasing device such as a pipe. In comparison, IIG \cite{iig} contains videos of real gas leaks, but it has only bounding box annotations and is captured on handheld cameras, which introduces substantial camera motion.

\begin{table*}[t]
    \centering
    \renewcommand{\arraystretch}{1.15}
    \begin{tabular}{l>{\centering\arraybackslash}p{2.5cm}>{\centering\arraybackslash}p{2.5cm}>{\centering\arraybackslash}p{2.5cm}>{\centering\arraybackslash}p{2.5cm}}
        \hline
        & \textbf{Gas-DB \cite{gasdb}}& \textbf{GasVid \cite{gasvid}}& \textbf{IIG \cite{iig}}&\textbf{Ours}\\
        \hline
        Format& Image Sequence& Video& Video&Video \\
  Video Clip Length& Short ($\sim$60 frames)& Long ($\sim$21K frames) &Medium ($\sim$500 frames)&Short-medium ($\sim$300 frames) \\
        Collecting Method & Controlled Release& Controlled Release& Real Emission &Computer Simulation \\
        Spatial Bias& Yes & Yes & No&No  \\
        Background & Complex & Simple & Complex&Complex  \\
        Ground Truth & Manually labeled Masks& Quantification Classes& Bounding Boxes&Priori Ground Truth  \\
        Scene Variation & 8 & 1 & 5&9\\
        Other Moving Objects & Yes & Rarely & Some&Yes  \\
        Frame Continuity & Low & High & High&High  \\
        Total Number of Frames & ~1.2k & ~700k & ~5k&~12k \\
        Inter-Video Similarity& Low & High & Low&Medium\\
        Availability & Public & Public & Upon Request&Public  \\
        \hline
    \end{tabular}
    \caption{Comparison of different datasets}
    \label{tab:dataset_comparison}
\end{table*}
We attempted to label pixel-level segmentation ground truth for GasVid \cite{gasvid}, but the semi-transparent, blurry boundaries of gas leaks made annotation difficult. Consequently, our proposed method of Priori Ground Truth—knowing the ground truth before generating input data—can address these challenges by establishing accurate annotations from the outset.

Datasets in all domains exhibit inherent biases. GasVid and Gas-DB, for example, have spatial biases related to specific releasing devices, and all three datasets can be influenced by factors such as camera type, location, and lighting. Compiling large-scale gas leak detection data is inherently difficult, resulting in relatively small datasets that risk propagating any underlying biases to trained models. Recently, zero-shot techniques have gained attention for their low implementation cost and independence from training data.

To advance this area of research, we propose a novel computer-synthetic dataset offering diverse leakage points, perfectly accurate segmentation ground truth, and stable video recordings containing multiple moving objects. We produce high-quality data that avoids human labelling and retains precise segmentations by overlaying realistically rendered leaks and interfering foreground elements onto varied backgrounds. In addition, we proposed a zero-shot method using a vision language model to avoid the model bias trained on this dataset. Our experiment shows that this method can achieve promising performance on this dataset and GasVid. 

Our contributions can be summarized as follows:
\begin{itemize}
\item We construct a diverse video-based computer-rendered dataset with complex backgrounds, interfering moving objects, and accurate ground truth.
\item We propose a new baseline algorithm that combines background subtraction and zero-shot object detection to segment gas leakage accurately.
\item We tested our method on a real-world dataset, GasVid \cite{gasvid}, and the results are promising. 
\end{itemize}

\section{Related Work}
\subsection{Gas Leak Detection and Datasets}
There are three main public datasets in the gas leak detection field. GasVid was proposed with GasNet \cite{gasvid} and VideoGasNet \cite{videogasnet}. It contains a video dataset of controlled gas release. Most videos include the sky as a background, with gas released from a chimney-like structure. However, it was originally used as a classification dataset to determine if there is leakage and the amount of leakage without localization or segmentation information. Segmentation is not only important for precise localization but also required for better quantification of the gas release \cite{gasformer}. Most of its videos also do not have interference from other moving objects such as humans or cars, making the background-subtraction method already able to segment out the foreground (leak) with decent performances. In GasNet \cite{gasvid} and VideoGasNet \cite{videogasnet}, authors used background subtraction to remove non-moving parts in the video and kept a ``soft" subtraction (without thresholding). Then, these subtracted frames (still frames in GasNet \cite{gasvid} and sequence of frames in VideoGasNet \cite{videogasnet}) were sent into a CNN or ConvLSTM \cite{convlstm} to classify if there is a leak in the frames and the amount of the leak. 

Gas-DB \cite{gasdb}, on the other hand, is a segmentation dataset. It contains over 1000 RGB-T images (images with 3 RGB channels and one thermal channel) with carefully labelled segmentation masks in different environments with other moving objects. The RGB channels provided more textual information than the thermal-only images. However, it was designed for image segmentation tasks without considering temporal information, which makes it hard to distinguish other similar-looking objects with leaks or hard-to-detect faint leakage. Although the images are collected in temporal sequence and can be connected as videos, the lengths are usually very short, and the frame continuity is very low. Their model uses cross-modality attention to leverage the information in both RGB channels and thermal channels, achieving 56.52\% IoU results. However, they split the dataset into training and validation sets using frame level instead of video level splits, meaning frames in the same video, which could have similar environments, can end up in training and validation sets. This means that when the model is applied to unseen environments, the performance could potentially drop. This is similar to the situation for background subtraction for seen scenes vs. unseen scenes (Section \ref{sec:bgs}).      
Additionally, both of these datasets have gas ``leaked" from the end of pipe- or chimney-like structures. Therefore, if we train a model on these raw images (i.e. not the background-subtracted images in GasNet\cite{gasvid} and VideoGasNet \cite{videogasnet}), the model could be biased toward these structures such that it will tend to relate these structures to the leakage, which is not necessarily the case in real scenarios. 

The Industrial Invisible Gas (IIG) dataset \cite{iig} is another recent IR-camera-captured dataset designed for object detection in real-world industrial environments. Unlike artificially simulated gas leaks, this dataset represents actual scenarios, avoiding biases associated with predefined leak locations, such as the ends of pipes. It consists of 5,569 images and includes five distinct scenes: pump oil seals, oil tank vents, gas stations, industrial chimneys, and other industrial settings. It was also captured as videos with high continuity, but the videos were captured using a handheld camera, which introduced camera jitter motion, making some video-based methods (such as background subtraction) hard to apply.

Furthermore, both GasVid \cite{gasvid} and GasDB \cite{gasdb} utilize real-world simulations with controlled gas releases. Although these methods aim to replicate real-world scenarios closely, they pose significant fire and explosion hazards (with the risk being lower in GasVid due to its open-air setting but higher in GasDB, where some experiments take place in partially enclosed spaces) and contribute to the release of greenhouse gases into the environment. 

According to the GasVid paper \cite{gasvid}, at least 12 kg of methane was released for the dataset used in the study—excluding emissions from testing and failed attempts (such as tank releases). While this quantity may be negligible globally, it still represents an avoidable environmental impact from a single experimental dataset. In contrast, a computer-generated dataset can achieve the same objectives without contributing to greenhouse gas emissions. In addition, humans' labelling of segmentation masks is inefficient and inaccurate. This is due to the gas plume's blurry boundary and transparent nature. 

The above-mentioned problems may be resolved using synthetic datasets, and synthetic datasets have been used in many areas. Guo \textit{et al.} \cite{Guo_Du_Shehata_2024} demonstrated that for molecular model captioning, using a rendered dataset provides a straightforward method to obtain large amounts of data without human involvement, significantly boosting the downstream tasks performance on real datasets. Wang \textit{et al.} \cite{M4SFWD_rendered_smoke1} used Unreal Engine 5 to simulate forest fire. Mao \textit{et al.} \cite{rendered_smoke_2} used 3D software to render forest fire smoke images and used CycleGAN \cite{cyclegan} to generate more images. Gu \textit{et al.} \cite{rendered_smoke3} also used the rendered dataset to simulate gas leakage to avoid manual labelling segmentation data; however, to the best of our knowledge, their dataset is not opened and is not clear whether it is video-based or image-based.

\subsection{Background Subtraction}
\label{sec:bgs}
Background subtraction (BGS) \cite{mog, mog2_2, knn_mog2_1, bsuv, bsuv2, ZBS} has been used to detect moving parts in a video. \footnote{Note that this is not the same as background removal, which is to segment the saliency objects in a single frame image to remove the background; an example use case is to blur or replace the background in online video conferences. For background removal, readers can refer to \cite{dis,u2net}} Non-deep-learning methods, such as MOG \cite{mog}, MOG2 \cite{mog2_2, knn_mog2_1}, and kNN \cite{knn_mog2_1}, do not require prior training or masks labelling for frames at the beginning of the video. 

Over an extended period, one class of supervised background subtraction methods is video or video-group-optimized deep learning methods \cite{lim_learning_2018} \cite{lim_foreground_2018_1}. These methods require some frames from the target video or target video group to be labelled (i.e., segmentation by humans of what objects are in the foreground) to perform well. When applied to unseen videos, their performance drops dramatically \cite{bsuv}. For example, the original F-score reported by FgSegNet v2 \cite{lim_learning_2018} using a video-optimized method was nearly perfect (0.9789), but when trained using the video-agnostic method by \cite{bsuv}, the f-score is only 0.3715.

BSUV-Net \cite{bsuv} is one of the first deep-learning-based methods that could be applied to unseen videos with good results. BSUV-Net 2.0 \cite{bsuv2} improved it by using stronger data augmentation. Other deep learning methods for unseen BGS include \cite{lin_foreground_2018, st-charles_subsense:_2015, 3dfr, zhao_universal_2022, yang_stpnet:_2022}.

Zero-shot background subtraction (ZBS) \cite{ZBS} introduced open vocabulary detection and segmentation for BGS in zero-shot settings. It uses an open vocabulary object detector to detect all objects in the image and uses their movement across the video to determine if the object is foreground or background. While ZBS demonstrates effectiveness in handling illumination changes and pixel noise, it presents a fundamental limitation: it requires objects to be detectable before background subtraction can occur. This requirement creates a significant problem for applications such as leak detection. In such scenarios, background subtraction is used as a preliminary step for object detection by extracting the moving objects that could be hard to detect on the original image—not the other way around. This creates a dependency loop for these applications.

\subsection{Vision Language Models}
Vision language models (VLMs) \cite{clip, gpt4, gpt4o, florence, llama3, llava, BLIP, glip, owlvit, owlv2, groundingdino, git} have been advanced dramatically recently and show promising results in tasks combining vision and language, especially in zero-shot settings, such as language guided classification, segmentation, object detection, and vision grounding. CLIP (Contrastive Language Image Pretraining) \cite{clip} is one of the earliest and foundational works in this field. It uses contrastive pre-text tasks where matched image-text pairs are used. This allows for zero-shot image classification by giving the model a list of candidate classes in natural language and calculating the similarity between the image and text features.
Other VLMs have expanded CLIP-like zero-shot capabilities to localization (including grounding and detection) \cite{groundingdino, improvinggrounding1, improvinggrounding2, owlv2, owlvit, florence}. By combining these localization models with the Segmente Anything Model (SAM) \cite{sam}, language-guided instance-level segmentation could be achieved, such as in Grounding SAM \cite{groundedsam} and APOVIS \cite{ma_apovis:_2025}. More details of these models can be found in the supplemental material.

\section{Dataset}
\begin{figure*}[h]
    \centering
    \includegraphics[width=\linewidth]{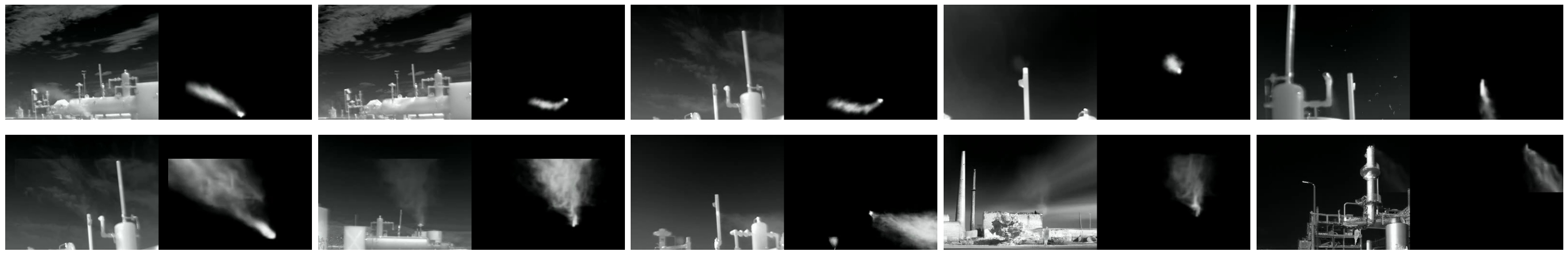}
    \caption{\textbf{Preview of our dataset.} These images are selected from 10 different videos. For each side-by-side subplot, the left one is the input frame, and the right one is the un-thresholded ground truth. Some of these videos use GasVid \cite{gasvid} as background, while others use DALL-E-2 \cite{dalle2} generated backgrounds.}
    \label{fig:examples}
\end{figure*}

We created the dataset by overlaying interfering foreground objects and gas leakage simulation footage onto background scenes. The foreground elements were sourced from two IR datasets, BU-TIV \cite{ir_dataset_wu} and CAMEL \cite{ir_dataset_saha, ir_dataset_gebhardt}, which include objects such as bats, cars, and humans. To extract objects of interest from these IR videos, we segmented the objects of interest using either thresholding or the SAM 2 model \cite{sam2} with box annotations from the original dataset. Background footage was selected from GasVid \cite{gasvid} from non-leak portions or generated using DALL-E-2 \cite{dalle2}.  GasVid backgrounds are used to ensure our dataset includes sensor noise, real-life lighting change, etc. DALL-E-2-generated backgrounds are used to diversify the background scenes. Gas leakage simulations were rendered in Blender using smoke simulation and force field. Some foreground objects, leakage simulations, and background scenes were reused in different combinations. For ground truth for segmentation, we used the generated ``smoke" footage at the same position as in the overlay. A detailed comparison of our dataset with GasVid and GasDB is provided in Table~\ref{tab:dataset_comparison}. 

Similar to Gas-Vid \cite{gasvid} and Gas-DB \cite{gasdb}, we removed some videos (26, 27, and 28) because they exhibit a strong, highly localized wind that causes smoke to disperse significantly.
We believed that such extreme, erratic wind behaviour is rare in real-life scenarios. Therefore, we decided to exclude these videos, as they do not represent typical conditions and introduce unrealistic challenges. We also removed video 24 because of a misalignment between the video and the ground truth. We have retained these videos in the published dataset so that readers can review them and assess our decision.

\section{Methods}
A pipeline of our method can be found in Figure \ref{fig:pipeline}, and the pseudo code can be found in supplemental material. 

\subsection{Background Subtraction and Enhancement}
A sequence of simulated IR frames is processed with background subtraction to extract the moving part of the video. We used a short history (30 frames) to avoid false positives from slow-moving objects such as clouds, which is the same approach used in GasNet \cite{gasvid}.

Instead of relying on built-in mask generation for different background subtraction (BGS) methods, we extracted the background image from the algorithm and then computed the absolute difference between the current frame and the background image:
\(I_{i}'=|I_{bg}-I_{i}|\)
where \(I_{i}'\) is the difference, \(I_{bg}\) is the background image obtained from the BGS algorithm, and \(I_{i}\) is the current frame.

Since the difference could be subtle, we enhanced the image by a factor \(\alpha\) and clipped the values between 0 and 255:
\(I_{i}'' = \min(\max(\alpha I_{i}', 0), 255)\).

Because the intensity of the difference may vary across different scenarios, we used an adaptive enhancement factor, as shown in Equation \ref{alpha}. We set the default factor to 15; however, this value could sometimes be too high when the intensity of the difference is large, leading to clipping and loss of image details. To mitigate this issue, we ensured that \(\mu_{I_{i}'} + \sigma_{I_{i}'}\) (one standard deviation above the mean) does not exceed 255 by selecting a lower \(\alpha\), as shown in Equation \ref{alpha}.
\begin{equation}
    \alpha = \min(\frac{255}{\mu_{I_{i}'} + \sigma_{I_{i}'}}, 15)
    \label{alpha}
\end{equation}

\begin{table*}[]
    \centering
    \begin{tabular}{c|c|>{\centering\arraybackslash}p{40mm}>{\centering\arraybackslash}p{40mm}>{\centering\arraybackslash}p{40mm}}
         BGS Method&    Refinement&Without Interference I/P/R/FLA& With Interference I/P/R/FLA&Overall I/P/R/FLA\\
    \hline
         Median& 
      Morph&0.50 / 0.63 / 0.74 / \textbf{0.85}& 0.30 / 0.52 / 0.44 / 0.68&0.43 / 0.59 / 0.63 / 0.79\\ 
 Median&  -&0.41 / 0.67 / 0.53 / \textbf{0.85}& 0.25 / 0.56 / 0.33 / 0.68&0.35 / 0.63 / 0.45 / \textbf{0.79}\\ 
 MOG2 \cite{mog2_2, knn_mog2_1}&  Morph&\textbf{0.56} / 0.67 / \textbf{0.8} / \textbf{0.85}& \textbf{0.38} / 0.56 / \textbf{0.57} / \textbf{0.69}&\textbf{0.5} / 0.63 / 0.7 / 0.79\\ 
 MOG2 \cite{mog2_2, knn_mog2_1}&  -&0.51 / \textbf{0.68} / 0.68 / \textbf{0.85}&0.35 / \textbf{0.6} / 0.47 / \textbf{0.69}&0.45 / \textbf{0.65} / 0.6 / \textbf{0.79}\\ 
 $k$-NN \cite{knn_mog2_1}&  Morph&0.23 / 0.36 / 0.46 / 0.78& 0.16 / 0.27 / 0.29 / 0.63&0.21 / 0.32 / 0.41 / 0.72\\ 
 $k$-NN \cite{knn_mog2_1}&  -&0.16 / 0.32 / 0.26 / 0.78& 0.11 / 0.24 / 0.18 / 0.63&0.14 / 0.29 / 0.23 / 0.72\\\end{tabular}
    \caption{\textbf{BGS Only Baseline on Our Dataset} We tested different background subtraction (BGS) methods with different refinement settings. We reported the Intersection over Union (IoU, I), precision (P), recall (R), and frame-level accuracy (FLA). Frame-level accuracy is the method's accuracy in classifying each frame to leak (with positive pixels) and no leak (all pixels are negative).}
    \label{tab:bgs_only}
\end{table*}
\begin{figure}[t]
    \centering
    \includegraphics[width=1\linewidth]{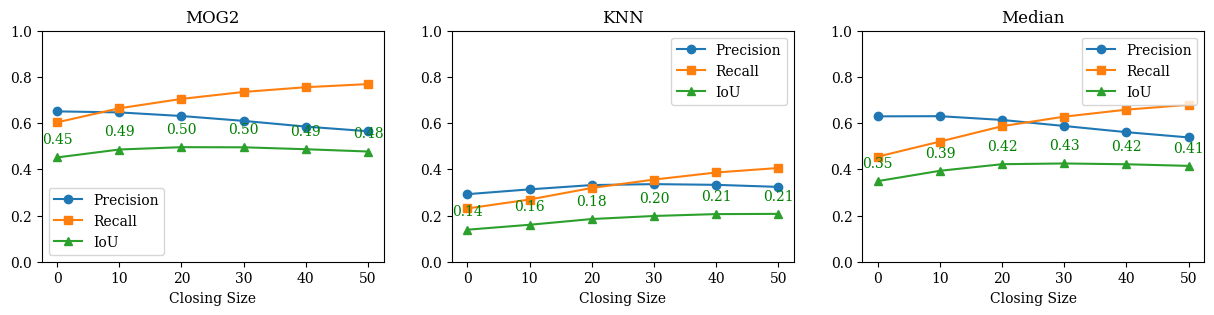}
    \caption{BGS-only baseline on our dataset with different morphological closing operation sizes }
    \label{fig:bgs_obly}
\end{figure}
\subsection{VLM Filtering}
After background subtraction, all moving objects, including non-leak objects like humans, cars, birds, etc, are also extracted. To select leaks, we leverage the zero-shot object detection capability of a Vision-Language Model (VLM) Owlv2 \cite{owlv2} to filter objects of interest (leak), with the VLM Threshold ($\tau_{VLM}$) as a hyperparameter that determines the threshold for a positive box output. However, using a prompt of ``gas leak" or something similar might not be ideal as these models are usually trained on RGB modality images, in which gas is usually non-visible. Nevertheless, gas leaks in IR images resemble steam or smoke in RGB images. Thus, we chose to use ``white steam" as the prompt for object detection. Detailed experiments with different prompts are shown in Section \ref{prompt_exp}. 
We also used one negative prompt (\textit{white human, car, bird, bike, and other objects}), such that when the objects were similar to ``white steam" but more similar to the negative prompt, the VLM could correctly avoid it to reduce false positives. Finally, we applied a non-maximum suppression based on the confidence of each bounding box and the IoU between them to reduce overlapping boxes.    

\subsection{Temporal Filtering}
Since the VLM only considers the current enhanced difference frame ($I{i}''$) as input, it lacks temporal information. This limitation can lead to transient false positives or false negatives, causing issues with poor segmentation and false alarms.

To address this problem, we assume that a leak does not appear or disappear suddenly. We implement a temporal filtering mechanism that ensures a detected box is considered valid only if, within the past $k_1$ frames, at least $n_1$ boxes have an IoU greater than $\tau_{tIoU_1}$ or absolute shift greater than $\tau_{tShift}$ with the current box. This prevents transient false positives caused by noise or non-leak objects. We used $k_1=10$, $n_1=1$, $\tau_{tIoU_1}=0.3$, and $\tau_{tShift}=40$ in our experiment. 

Similarly, we assume that a leak will not vanish suddenly. Therefore, if no leak is detected in the current frame, we look over the past \(k_2\) frames and compare all detected boxes across these frames. If two boxes in different frames have an IoU greater than \(\tau_{tIoU_2}\), we infer that the leak is still present and add the corresponding box to the current frame. We used $k_2=3$ and \(\tau_{tIoU_2}=0.3\) in our experiment. These hyperparameters were not tuned extensively to avoid overfitting to a certain dataset. 

\begin{figure}
    \centering
    \includegraphics[width=0.5\linewidth]{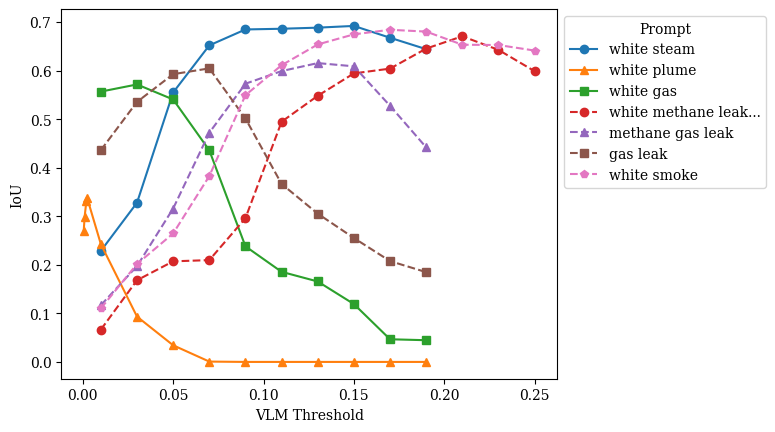}
    \caption{\textbf{Performance of VLM with different prompts and thresholds}}
    \label{fig:prompt_t_vlm}
\end{figure}

These two filters provide a simple method to balance response time against false positive and false negative rates, leveraging the assumption that leaks are generally continuous. A detailed pseudocode implementation is provided in Algorithm 1 in the supplemental material, and we demonstrate its effectiveness in the experiment section.
\subsection{Segmentation}
After temporal filtering, the boxes are passed to SAM 2 \cite{sam2} to generate segmentation masks. SAM 2 produces a mask for each given box, which is combined using an OR operator. We utilize SAM 2 because it is less susceptible to noise and can effectively disregard non-primary objects, ensuring more accurate segmentation of the target subject.

\begin{table*}
    \centering
    \begin{tabular}{c>{\centering\arraybackslash}p{1cm}>{\centering\arraybackslash}p{1cm}>{\centering\arraybackslash}p{12mm}>{\centering\arraybackslash}p{8mm}|>{\centering\arraybackslash}p{35mm}>{\centering\arraybackslash}p{35mm}|>{\centering\arraybackslash}p{35mm}}
        \hline
        BGS & VLM Filtering & Temporal Filtering & Seg. &  $\tau_{VLM}$&Without Interference I/P/R/FLA& With Interference I/P/R/FLA&  Overall I/P/R/FLA\\
        \hline
        \checkmark &               &               &               None&  -&0.56 / 0.64 / 0.83 / 0.85& 0.38 / 0.53 / 0.58 / 0.69& 0.5 / 0.61 / 0.73 / 0.79\\
        \checkmark & \checkmark    &               & SAM 2&  0.09&0.67 / 0.81 / 0.79 / \textbf{0.88}& 0.54 / 0.79 / 0.65 / 0.83& 0.62 / 0.80 / 0.74 / 0.86\\
                   & \checkmark    & \checkmark    & SAM 2&  0.19&0.22 / 0.39 / 0.28 / 0.57& 0.46 / 0.65 / 0.59 / 0.74& 0.31 / 0.49 / 0.4 / 0.63\\
 \checkmark    & \checkmark    & \checkmark    & Trad.&  0.12&0.57 / \textbf{0.85} / 0.65 / 0.83& 0.35 / \textbf{0.88} / 0.37 / 0.72&0.49 / \textbf{0.86} / 0.55 / 0.79\\
 \checkmark & \checkmark & \checkmark & SAM 2&  0.12&\textbf{0.70} / 0.83 / \textbf{0.82} / 0.87& \textbf{0.69} / 0.79 / \textbf{0.84} / \textbf{0.92}&\textbf{0.69} / 0.82 / \textbf{0.82 }/ \textbf{0.89}\\
        \hline
 
    \end{tabular}
    \caption{\textbf{Ablation study of different components with IoU (I), Precision (P), Recall (R), and Frame-Level Accuracy (FLA).} In the segmentation column (Seg.), traditional (Trad.) means Otsu \cite{otsu} combined with morphological transformations. This analysis corresponds to our ablation study, detailed in Section~\ref{sec:ablation}.}
    
    \label{tab:ablation}
\end{table*}

\begin{table*}[]
    \centering
    \begin{tabular}{c|ccc}
    \hline
         Prompt&  Without Interference I/P/R/FLA &With Interference I/P/R/FLA &Overall I/P/R/FLA\\ \hline
         white gas& 
     0.59 / 0.82 / 0.70 / 0.88& 0.55 / 0.74 / 0.66 / 0.80&0.57 / 0.80 / 0.67 / 0.83\\
 white plume& 0.35 / 0.55 / 0.52 / 0.67& 0.31 / 0.48 / 0.45 / 0.63&0.34 / 0.52 / 0.50 / 0.66\\
 white steam& \textbf{0.71} / 0.82 / \textbf{0.84} / 0.90& \textbf{0.70} / \textbf{0.83} / \textbf{0.82} / \textbf{0.91}&\textbf{0.69} / \textbf{0.83} / 0.81 / 0.88\\
 white methane leak...& 0.70 / 0.82 / 0.83 / 0.89& 0.62 / 0.75 / 0.77 / 0.89&0.67 / 0.79 / 0.81 / 0.89\\
 methane gas leak& 0.62 / 0.79 / 0.75 / 0.82& 0.63 / 0.77 / 0.77 / 0.87&0.62 / 0.75 / 0.79 / 0.86\\
 gas leak& 0.62 / 0.79 / 0.76 / 0.83& 0.57 / 0.75 / 0.70 / 0.86&0.60 / 0.77 / 0.74 / 0.84\\
 white smoke& \textbf{0.71} / \textbf{0.83} / \textbf{0.84} / \textbf{0.91}& 0.65 / 0.79 / 0.79 / \textbf{0.91}&0.68 / 0.81 / \textbf{0.82} / \textbf{0.91}\\ \hline
 \end{tabular}
    \caption{\textbf{Performance of different prompts on without interference, with interference, and overall.} The complete form of ``The white methane leak..." is ``white methane leak on black background in the infrared image." The prompts containing ``white smoke" and ``white steam" yielded the highest performance. In terms of overall performance, as measured by Intersection over Union (IoU), the prompt with ``white steam" demonstrated a slight advantage over the prompt with ``white smoke". }
    \label{tab:my_label}
\end{table*}

\section{Experiments and Results}
\subsection{Settings}
In this paper, for each method tested, we performed a hyper-parameter test on key parameters such as morphological kernel size and threshold. We ensured the approach aligns with real-world frame rate limitations—where hardware performance is constrained despite the need for real-time monitoring—and acknowledge that closely spaced frames are often similar, making individual evaluation unnecessary. We process every frame using the BGS algorithm but only perform VLM filtering and subsequent stages (including evaluation) every 5 frames. To ensure consistent comparison for the BGS-only baseline, we also performed BGS for every frame but only evaluated the result every 5 frames. The Owl-V2 model is loaded from Hugging Face using 4bit quantization and float16 computation type. The SAM 2 model is 2.1 Hiera Small using an official repository. 
We reported four metrics for comparison: IoU (I), precision (P), recall (R), and frame-level accuracy (FLA). Frame-level accuracy is the accuracy of frame-level classification. If any pixels in the frame are segmented, it is considered positive. IoU is aggregated over a whole video, and then all videos' IoU are averaged to the final reported value. Other ways to calculate the IoU would result in different values. 

\subsection{BGS-Only Baseline}
\label{sec:bgs_only}
In GasNet \cite{gasvid} and VideoGasNet \cite{videogasnet}, employing only background subtraction (BGS) produces a clear and accurate segmentation of leakage (foreground) due to the static background and the absence of interfering moving objects such as cars or people. Therefore, we established a baseline using only BGS on our dataset.

We tested several BGS methods, as detailed in Table \ref{tab:bgs_only}. Each method was executed with a history length of 30 frames. The background image, extracted from the background model, is subtracted from the current frame; the resulting difference image is then scaled by a factor of 15 and thresholded at 40. These hyperparameters were manually tuned using the MOG method and then uniformly applied across all tested methods, as exhaustive hyperparameter tuning for each individual method was impractical.

To enhance segmentation masks obtained from BGS methods, we investigated the impact of morphological operations. Specifically, we applied morphological opening to remove salt noise and closing to merge fragmented segments arising from weak leakage signals. Closing kernel sizes between 10 and 50 were tested, with the optimal results displayed in Table \ref{tab:bgs_only}\footnote{Reported best results correspond to the highest Intersection over Union (IoU) settings, where precision, recall, and frame-level accuracy (FLA) are also provided.}. Morphological opening was consistently applied in all tests, while morphological closing was selectively applied only when indicated by the "morph" setting.

Results are separately reported for scenarios with no interfering objects and those including moving objects. The performance of each method with varying morphological closing kernel sizes is shown in Figure \ref{fig:bgs_obly}. From Figure \ref{fig:bgs_obly} and Table \ref{tab:bgs_only}, we observe that MOG2 achieves the best overall performance, and larger morphological kernel sizes generally produce improved results.

For the pseudo code of the BGS-Only Baseline, please see supplemental material.

\subsection{Ablation Study} \label{sec:ablation}
\begin{figure}
    \centering
    \includegraphics[width=0.3\linewidth]{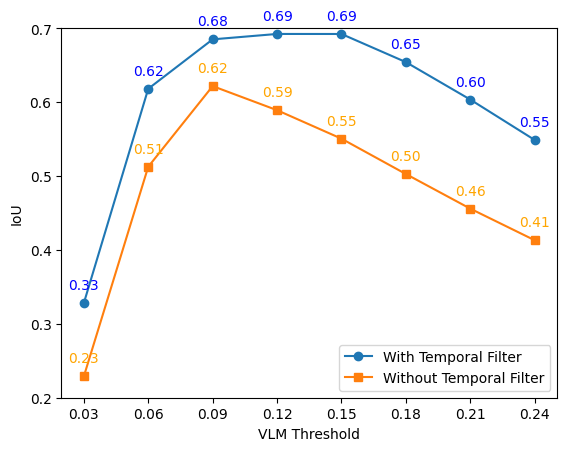}
        \caption{\textbf{Performance of Method with and without Temporal Filter across Different VLM Thresholds ($\tau_{VLM}$)}}
    \label{fig:vlm_threshold_filter}
\end{figure}

Our method consists of four main components: background subtraction, VLM filtering, temporal filtering, and SAM 2 \cite{sam2} for segmentation. To systematically investigate the contribution and effectiveness of each individual component in our pipeline, we conducted an ablation study as summarized in Table~\ref{tab:ablation}. Specifically, we compared five experimental cases:

\begin{enumerate}
    \item \textbf{BGS-Only Baseline}: This row represents the best-performing result from the background-subtraction-only experiments described in Section~\ref{sec:bgs_only}. This could be considered an improved version of the first step of GasNet \cite{gasvid} and VideoGasNet \cite{videogasnet}.
    
    \item \textbf{BGS + VLM Filtering + SAM 2 (optimal threshold, without temporal filtering)}: In this condition, we integrated visual-language model (VLM) filtering into the best-performing background subtraction system and employed Segment Anything Model 2 (SAM 2)~\cite{sam2} for converting bounding boxes generated by the VLM into segmentation masks. Noticeably, the only distinction from our complete proposed method is the absence of temporal filtering. By comparing this condition with our complete method, we analyzed the incremental value provided by the temporal filtering component.  
       
    Note that the optimal threshold for the VLM filtering differs between this setting and our proposed method; thus, we conducted separate threshold sweeps for both versions. The corresponding results are visualized in Figure~\ref{fig:vlm_threshold_filter}. From this figure, we also observed that our complete method has a broader effective threshold range, indicating higher robustness against threshold variations. 
    
    \item \textbf{Proposed Method without Background Subtraction (VLM + temporal filtering + SAM 2, no BGS)}: In this configuration, background subtraction was omitted. The hypothesis was that not eliminating stationary objects would lead to difficulties in both correctly identifying leaks and avoiding false positives from non-leak objects. To achieve the best possible performance under this configuration (since its modality is different from our configuration), a grid search was conducted on the enhancement factor and VLM threshold to determine the optimal settings. Details of the grid search are provided in the supplementary material.
    
    \item \textbf{Proposed Method with Traditional Segmentation (BGS + VLM filtering + temporal filtering + Otsu~\cite{otsu})}: We replaced the powerful SAM 2 segmentation method with the conventional segmentation technique proposed by Otsu~\cite{otsu}, combined with simple morphological operations. Given that our test scenario consists primarily of a clear white leakage region against a dark background, it might be possible to achieve reasonable segmentation results using simpler methods. Thus, we evaluated this setting to establish the necessity and advantage of employing the more advanced SAM 2 segmentation algorithm.  
    
    \item \textbf{Complete Proposed Method (BGS + VLM filtering + temporal filtering + SAM 2)}: This condition represents our complete proposed approach, integrating all the discussed components to achieve robust leakage detection and segmentation.
\end{enumerate}

The ablation study shows that using only background subtraction (case 1) provides reasonable results, especially in cases where there is no With Interference. However, when VLM filtering and Segment Anything Model 2 are added (case 2) for segmentation, performance improves significantly, with an overall IoU increase of more than 10\%. This enhancement allows for better filtering of non-leak objects.

In contrast, removing background subtraction (case 3) to detect moving objects leads to the worst performance, even lower than the baseline of case 1. Without background subtraction, the model struggles to identify leaks and correctly classify non-leak objects, as indicated by both low precision and recall.
Case 4 setup performed similarly to the background subtraction baseline but with a high precision and low recall. Upon visual inspection of a few images, we noted that some bounding boxes failed to encompass objects fully. Traditional segmentation could not extend the segmentation masks beyond bounding box limitations, whereas SAM 2 can use semantic information to capture entire objects.

With the full model (case 5), performance reaches 69\% IoU, showing an almost 20\% improvement over the MOG baseline and approximately a 40\% boost compared to using only a visual language model, temporal filtering, and segmentation. Comparing this with case 2, where no temporal filtering is used, we achieved a 7\% IoU boost in overall cases and a 15\% increase in With Interference. These two cases have a similar precision, but case 5 has a higher recall and higher frame-level accuracy. This could be due to the propagation mechanism in the filter. By comparing the performances of case 2 and case 5 across different VLM Thresholds in Figure \ref{fig:vlm_threshold_filter}, we can also observe that with temporal filtering, the method is more robust to changes in the threshold. 


\subsection{Prompt Comparison}
\label{prompt_exp}

We hypothesized that the model might struggle to understand abstract or uncommon concepts such as ``gas", ``methane," and ``leak." To address this, we used ``white steam" as a positive prompt in the object detection phase. To test our hypothesis, we experimented with different prompts. Since different prompts may have different optimal values $\tau_{VLM}$, we performed a sweep across $\tau_{VLM}$ for each prompt. The results are shown in Figure \ref{fig:prompt_t_vlm}.

In our comparison, ``white steam" and ``white smoke" achieved the best performance, which we attribute to their frequent occurrence in natural language. Additionally, the prompt ``white methane leak ..." also performed well, likely due to its detailed description specifying the infrared modality and black background. However, ``white gas" exhibited poor performance, which we hypothesize is because ``gas" is generally invisible in RGB images, which is what most VLMs are trained on. For a more detailed analysis, please refer to the supplemental material.

\subsection{Qualitative Experiments on GasVid}
We conducted qualitative experiments on GasVid to evaluate the method on real-world videos. Due to the lack of high-quality mask annotations, results were assessed visually. Visualizations are available in the GitHub repository. In the absence of leakage, the model generally avoided frame-level false positives. In frames containing leakage, the model could misclassify parts of the foreground as positive pixels because of background subtraction artifacts. The overall performance remains acceptable, with the primary issue being the over-segmentation of foreground objects. Small leaks sometimes led to false negatives. Please check the repository preview for more information.

\section{Conclusion and Furture Work}
In this work, we introduced a synthetic dataset with diverse backgrounds, interfering foreground objects, and precise segmentation labels for gas leak detection. Our zero-shot method leverages this dataset to significantly improve segmentation performance, achieving an IoU of 69\% and outperforming baseline methods based solely on background subtraction or zero-shot object detection. Unlike previous end-to-end approaches requiring labeled data, our method (LangGas) eliminates this need, reducing potential bias and labor-intensive labeling. However, it operates at only 2-3 FPS on an Nvidia A6000 GPU and 1 FPS on Quadro RTX 3000 (a low-end GPU), excluding skipped frames. This is still sufficient for industrial monitoring applications but leaves room for improvement. Potential extensions include fire detection, asteroid or supernova detection, camouflage detection, and microorganism tracking. Replacing background subtraction with optical flow could also enable use in scenarios with moving cameras, such as the IIG \cite{iig} dataset. Although quantitative comparisons on the GasVid dataset are limited due to a lack of ground truth, qualitative results (available on GitHub) show promising real-world performance, suggesting practical applicability.
\section*{Acknowledgment}
This research is supported by NFRF GR024473 and CFI GR024801.
\newpage
{
    \small
    \bibliographystyle{ieeenat_fullname}
    \bibliography{main}
}

\clearpage
\setcounter{page}{1}
\maketitlesupplementary
\section{Methane Release From GasVid}
The total amount of methane released during the capture of the GasVid dataset can be calculated using Equation \ref{methane}, where $m_{\text{total}}$ represents the total mass of methane released, $n$ is the number of videos, $i$ denotes the class label corresponding to the flow rate, and $m_i$ is the flow rate of the $i$-th class in g/h. Each flow rate lasted for 3 minutes. Based on the flow rate data from the GasVid paper, the total methane release is 12906.385g.

\begin{equation}
    \label{methane}
    \begin{aligned}
        m_{\text{total}} &= n \times \sum_{i=0}^{7} \left( \frac{m_i}{60} \times 3 \right) \\
        &= \frac{31 \times 3}{60} \times \sum_{i=0}^{7} m_i \\
        &= 1.55 \times 8326.7 \text{ g} \\
        &= 12906.385 \text{ g}
    \end{aligned}
\end{equation}

\begin{algorithm}
\label{temporal_filtering_alg}
\caption{Temporal Filtering Algorithm}
\label{tfilter}
\KwIn{\textit{current boxes}, \textit{past boxes}, \textit{image size}}
\KwOut{\textit{valid boxes}}

Set \textit{valid boxes} as an empty list\;

\For{each \textit{current box} in \textit{current boxes}}{
    \If{\textit{area of current box} $>$ \textit{image area} $\times$ \textit{ignore large threshold}}{
        Skip this box\;
    }
    
    Set \textit{matched boxes} to 0\;
    
    \For{each \textit{past frame boxes} in the last \textit{maximum past frames}}{
        Find overlap between \textit{current box} and \textit{past frame boxes}\;
        Find position difference between \textit{current box} and \textit{past frame boxes}\;
        
        \If{any overlap $>$ \textit{IoU match threshold} OR all position differences $<$ \textit{absolute shift threshold}}{
            Increase \textit{matched boxes} by 1\;
        }
    }
    
    \If{\textit{matched boxes} $\geq$ \textit{match threshold}}{
        Add \textit{current box} to \textit{valid boxes}\;
    }
}

Set \textit{matched boxes} as an empty list\;

\If{\textit{valid boxes} is empty AND \textit{past boxes} length > 3}{
    \For{each \textit{first frame} in the last 3 frames}{
        \For{each \textit{second frame} in the last 3 frames}{
            \If{\textit{first frame} == \textit{second frame}}{
                Skip this frame\;
            }
            
            \For{each \textit{box in first frame}}{
                \If{any overlap with \textit{boxes in second frame} $>$ \textit{IoU match threshold}}{
                    Add \textit{box in first frame} to \textit{matched boxes}\;
                }
            }
        }
    }
}

\Return \textit{valid boxes}\;

\end{algorithm}

\section{Further Review of VLMs}

\begin{figure*}
    \centering
    \includegraphics[width=1\linewidth]{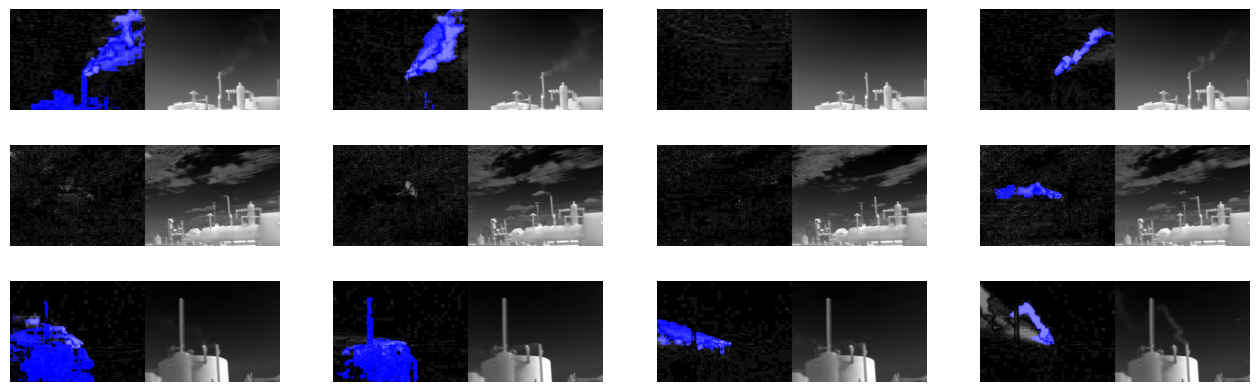}
   \caption{\textbf{Selected Samples from GasVid}: The two left columns display failure cases, and the two right columns show successful cases. In each pair, the left image shows the background subtraction result, with blue indicating the segmentation output (artifacts may appear), while the right image is the original frame. The three rows correspond to videos with GasVid IDs 1239, 2570, and 2579, recorded at distances of 12.6m, 15.6m, and 6.9m, respectively.}
    \label{fig:gasvid_sample}
\end{figure*}
CLIP \cite{clip} is a remarkable work that has inspired downstream work in segmentation, detection, etc. LSeg \cite{lseg} adapted CLIP into segmentation by calculating the similarity of the text query with every pixel on the feature map of the image, classifying each pixel into one of the text queries. It then used a special regulation block to decode the feature map into segmentations. This straightforward way has also been used in OwlVit \cite{owlvit} and Owl-V2 \cite{owlv2}. In OwlVit, they pre-trained the CLIP encoder using contrastive loss and transferred the model into detection by removing the pooling operation with a classification and localization head to archive language-guided detection. 

Besides the image-text contrastive loss function, align before fuse (ALBEF) \cite{albef} also used image-text matching and masked language modelling like in BERT \cite{devlin_bert:_2019}. Their model has an image encoder, a text encoder, and a multimodality encoder. 

Although ALBEF was not trained on grounding or localization tasks, their Grad-CAM \cite{selvaraju_grad-cam:_2019} has shown a strong localization correlation between phrases and text. This is further improved by \cite{improvinggrounding1, improvinggrounding2}. Grounding-DINO \cite{groundingdino} and GLIP \cite{glip}, on the other hand, are specifically trained on grounding tasks and trained in object detection fashion by producing bounding boxes for phases. Both the Grad-CAM and the bounding box can be used to prompt a segmentation model such as SAM \cite{sam}, or SAM 2 \cite{sam2} to generate language-guided instance segmentation masks like in Grounding-SAM \cite{groundedsam} and APOVIS \cite{ma_apovis:_2025}.

Another line of work took a generative approach \cite{gpt4, gpt4o, llama3, llava, florence, lai_lisa:_2024, wang_llm-seg:_2024, bai_one_2024}. In these works, GPT-4 serials \cite{gpt4, gpt4o} and llama-like \cite{llama3, llava} models use pure language as an interface, take in instruction as text prompt and generate output as pure text (such as location information in coordination). Florence, on the other hand, uses special tokens for different tasks (such as segmentation, detection, etc) and also uses special tokens for generated results. Some other works \cite{lai_lisa:_2024, wang_llm-seg:_2024, bai_one_2024} also used special tokens for segmentation results. 


\section{Qualitative Experiments on GasVid}
We excluded videos recorded at 18.6 m (following VideoGasNet \cite{videogasnet}) and selected examples showing two failure cases and two successful cases, as shown in Figure \ref{fig:gasvid_sample}. The experiment used MOG2 as the background subtractor, OWLv2 \cite{owlv2} as the visual language model with a threshold of 0.06, enhancement factor of 10, and both temporal filtering and SAM 2 enabled. The results indicate that the model can localize and segment leakage with reasonable performance, although worse than the synthetic dataset due to real-world noise, artifacts in background subtraction, etc. Future work should be done on how to improve this method on real-world captured videos. 

In the success cases, two samples (from the third column and first two rows) are true negatives, showing that noise is not mistakenly segmented as a leak, while the remaining examples are true positives with well-aligned segmentation boundaries. In the sample in the fourth column of the third row, the model avoids an artifact from background subtraction that is not a leak. In the failure cases, the first and third videos show over-segmentation of non-leak objects, and in the second video, the leak is missed (false negative) due to the larger distance. We provided 4 full video results in the attached video. 

\begin{figure}
    \centering
    \includegraphics[width=\linewidth]{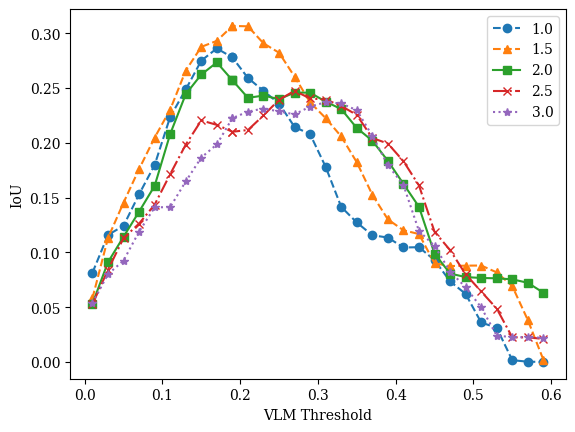}
    \caption{\textbf{Grid Search For Configuration without Background Subtraction:} We did a grid search on the enhancement factor and VLM threshold for the configuration without background subtraction. Different lines show different enhancement factors. The best performing point is when the enhancement factor is 1.5 and the VLM Threshold is 0.19. Results in this setting are values reported in Table \ref{tab:ablation}.}
    \label{fig:grid}
\end{figure}

\section{Prompts Comparison}
In our study on different prompts, ``white steam" and ``white smoke" performed the best, whereas ``white plume" exhibited the worst performance. We hypothesize that the superior performance of ``white steam" and ``white smoke" is due to their explicit description of both the substance (smoke or steam) and its colour (white). In contrast, the poor performance of ``white plume" is likely because ``plume" is a relatively uncommon word.  

Notably, the prompts ``white gas" and ``gas leak" also performed poorly. We attribute this to the fact that, in the training data of vision-language models (VLMs), ``gas" is often associated with ``gas station" rather than referring solely to a gaseous substance. As a result, the model may tend to link ``gas" to ``gas station" or ``gas stove," leading to suboptimal performance. Additionally, since gases are generally invisible in RGB images, and RGB is likely the primary modality in the training dataset, the model may struggle to associate the term ``gas" with its visual characteristics in infrared imagery. This suggests that the poor performance of prompts containing ``gas" is likely due to a mismatch between the term's associations in the training data and its expected visual representation in real-world scenarios.  

Another notable observation is that the long prompt, ``white methane leak on black background in the infrared image," achieved near-optimal performance, only slightly worse than the best-performing prompts. We hypothesize that while the VLM may not have a strong understanding of ``methane," the explicit description of the black background and the infrared image modality provide sufficient context for the model to generate accurate outputs.
\begin{algorithm}
\label{alg:bgs_only}
\caption{Background Subtraction (BGS) with Morphological Operations}
\SetAlgoLined
\SetNlSty{}{}{:}
\KwIn{Video frames $I_t$, history length $H=30$, scaling factor $s=15$, threshold $T=40$, kernel sizes for opening and closing}
\KwOut{Segmentation masks $D_t$}
Initialize background model with $H$ previous frames\;
\For{each frame $I_t$}{
Compute background model $B_t$\;
Compute difference image $D_t = |I_t - B_t|$\;
Apply scaling: $D_t \gets D_t \times s$\;
Threshold: $D_t \gets (D_t > T)$\;
Apply morphological opening on $D_t$ (remove salt noise)\;
\If{morphological closing enabled}{
Apply morphological closing on $D_t$ (merge segments)\;
}
}
\end{algorithm}

\begin{algorithm}
\label{alg:proposed}
\caption{Proposed IR Gas Leak Detection Method}
\KwIn{IR video sequence $\{I_i\}_{i=1}^{N}$}
\KwOut{Segmentation masks for gas leaks}

Initialize background subtraction method (e.g., MOG2)\;
Set $Prompt \gets$ ``white steam''\;
Set $NegativePrompt \gets$ ``white human, car, bird, bike''\;
Set VLM threshold $\tau_{VLM}$\;
Set $history \gets \{\}$\;
\For{each frame $I_i$ in the sequence}{
    Extract background image: $I_{bg} \gets$ BGS($I_i$)\;
    Compute absolute difference: $I_i' \gets |I_{bg} - I_i|$\;
    Compute enhancement factor: $\alpha \gets \min\left(\frac{255}{\mu_{I_i'} + \sigma_{I_i'}}, 15\right)$\;
    Enhance image: $I_i'' \gets \text{clip}(\alpha \cdot I_i', 0, 255)$\;
    
    $Boxes_i \gets$ VLM($I_i''$, $Prompt$, $NegativePrompt$, $\tau_{VLM}$)\;
    $Boxes_i \gets$ TemporalFiltering($Boxes_i$, $history$, size($I_i$))\;
    $history$ = $history$ + $\{Boxes_i \}$\;
    \If{$size(history) > 10$}
    {
        $history$.pop(0)\;
    }
    
    Obtain masks from $Boxes_i$ using SAM 2\;
    Combine all masks with OR operation to form final mask for frame $i$\;
}
\Return Segmentation masks for sequence\;
\end{algorithm}

\end{document}